\relax  
\documentclass[letterpaper]{article}
\usepackage{times}
\usepackage{helvet}
\usepackage{courier}

\usepackage[ margin=1.2in]{geometry}

\usepackage[utf8]{inputenc} 
\usepackage[T1]{fontenc}    
\usepackage{url}            
\usepackage{booktabs}       
\usepackage{amsfonts}       
\usepackage{nicefrac}       
\usepackage{microtype}      
\usepackage{graphicx}
\usepackage{amsmath}

\usepackage{subcaption}

\usepackage[svgnames]{xcolor}
\usepackage{xcolor,colortbl} 
\newcommand{\R}{\mathbb{R}}  
\providecommand{\scal}[2]{\left\langle{#1},{#2}\right\rangle}    

\begin{document}
%
\title{Bridging the Gaps Between Residual Learning, Recurrent Neural Networks and Visual Cortex}

\author{
Qianli Liao  and Tomaso Poggio\\     
Center for Brains, Minds, and Machines, McGovern Institute for Brain Research, \\ 
Massachusetts Institute of Technology, Cambridge, MA, 02139. 
}

\date{Sept. 15, 2016 \footnote{Written in 2016, released as it is.}}    

\maketitle

\begin{abstract}
  We discuss relations between Residual Networks (ResNet), Recurrent
  Neural Networks (RNNs) and the primate visual cortex. We begin with
  the observation that a special type of shallow RNN is exactly
  equivalent to a very deep ResNet with weight sharing among the
  layers. A direct implementation of such a RNN, although having
  orders of magnitude fewer parameters, leads to a performance similar
  to the corresponding ResNet. We propose 1) a generalization of both
  RNN and ResNet architectures and 2) the conjecture that a class of
  moderately deep RNNs is a biologically-plausible model of the
  ventral stream in visual cortex. We demonstrate the effectiveness of
  the architectures by testing them on the CIFAR-10 and ImageNet
  dataset. \footnote{ Pretrained model on ImageNet is available in \url{https://gitlab.com/liaoq/biological_rnn_resnet}  }             
\end{abstract}

\section{Introduction}
Residual learning \cite{he2015deep}, a novel deep learning scheme
characterized by ultra-deep architectures has
recently achieved state-of-the-art performance on several popular
vision benchmarks. The most recent incarnation of this idea
\cite{he2016identity} with hundreds of layers demonstrate consistent
performance improvement over shallower networks. The 3.57\% top-5 error
achieved by residual networks on the ImageNet test set arguably rivals
human performance.
 
Because of recent claims \cite{YaminsDicarlo2016} that networks of the
AlexNet\cite{krizhevsky2012imagenet} type successfully predict properties of neurons in
visual cortex,  one natural question arises: how similar is an ultra-deep residual
network to the primate cortex? A notable difference is
the depth. While a residual network has as many as 1202
layers\cite{he2015deep}, biological systems seem to have one or two orders of
magnitude less, if we make the customary assumption that a weighted linear combination layer (plus a nonlinearity) in the NN
architecture corresponds to a cortical area. In fact, there are about half a dozen
areas in the ventral stream of visual cortex from the retina
to the Inferior Temporal cortex. Notice that it takes in the
order of 10ms for neural activity to propagate from one neuron to
another one (remember that spiking activity of cortical
neurons is usually well below 100 Hz). The evolutionary advantage of
having fewer layers is apparent: it supports rapid (100msec from image
onset to meaningful information in IT neural population) visual recognition,
which is a key ability of human and non-human primates
\cite{thorpe1996speed,Serre2007}.

It is intriguingly possible to account for this discrepancy by taking
into account recurrent connections within each visual area. Areas in
visual cortex comprise six different layers with lateral and feedback
connections \cite{lamme1998feedforward}, which are believed to mediate
some attentional effects
\cite{buchel1997modulation,lamme1998feedforward,ito1999attention,rao1999predictive,hupe1998cortical}
and even learning (such as backpropagation \cite{liao2015important}).
``Unrolling'' in time the recurrent computations carried out by the visual
cortex provides an equivalent ``ultra-deep'' feedforward network, which
might represent a more appropriate comparison with the state-of-the-art
computer vision models.

In addition, we conjecture that the effectiveness of recent
``ultra-deep'' neural networks primarily come from the fact they can
efficiently model the recurrent computations that are required by the
recognition task. We show evidence for this conjecture by
demonstrating that 1. a deep residual network is formally equivalent
to a specific shallow RNN \footnote{We use the term ``RNN'' to broadly refer to any neural network with recurrent activations, not the ``vanilla/plain RNN'' (the baseline model people often use to compare with LSTMs). }; 2. such a RNN with weight sharing, thus with orders 
of magnitude less parameters (depending on the unrolling depth), can
retain most of the performance of the corresponding deep residual
network.
 
Furthermore, we generalize such a RNN to a class of more biologically
plausible models of visual cortex to account for multi-stage recurrent
processing and show their effectiveness on the CIFAR-10 and ImageNet
dataset. 

Another minor contribution of this work is that, we propose
using \textit{time-specific} batch normalization (TSBN, defined in
Section \ref{sec:bn}) in hidden-to-hidden transitions of RNN. We are
also the first to show when TSBN is used, there is no difficulty
training RNNs (even multi-state fully recurrent ones) with ReLUs. This
was previously believed to be difficult without careful weight
initializations \cite{le2015simple}. See supplementary materials for
additional experiments on character-level language modeling using our
model.

\section{Equivalence of ResNet and a specific RNN}
\subsection{Intuition}
\begin{figure}[h]
  \renewcommand\figurename{\small Figure}  
\begin{center}
  \includegraphics[width=\linewidth]{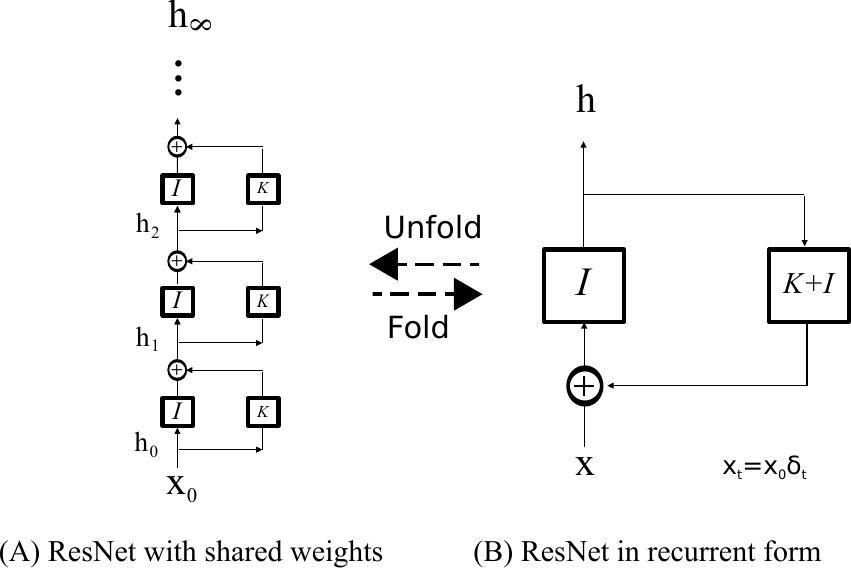} 
\end{center}
\caption{\small A formal equivalence of a ResNet (A) with weight sharing and 
  a RNN (B). I is the identity operator. K is an operator denoting the
  nonlinear transformation called $f$ in the main text. $x_t$ is the
  value of the input at time $t$. $\delta_t$ is a Kronecker delta function.}
\label{fig:feedback_system}
\end{figure}

We discuss here a very simple observation: a Residual Network (ResNet)
approximates a specific, standard Recurrent Neural Network (RNN)
implementing the discrete dynamical system described by 
\begin{equation}
  h_t = K \circ (h_{t-1}) + h_{t-1}
\end{equation}

where $h_t$ is the activity of the neural layer at time $t$ and $K$ is
a nonlinear operator. Such a dynamical systems corresponds to the
feedback system of Figure \ref{fig:feedback_system} (B). Figure
\ref{fig:feedback_system} (A) shows that unrolling in (discrete) time
the feedback system gives a deep residual network with the same (that
is, shared) weights among the layers. The number of layers in the
unrolled network corresponds to the discrete time iterations of the
dynamical system. The identity shortcut mapping that characterizes
residual learning appears in the figure.

From a biological perspective, a neuron's current activation,
characterized by Excitatory Postsynaptic Potential (EPSP) of soma
(i.e., activation before nonlinearity) is roughly the sum of inputs
plus the residual EPSP from the previous time step. The latter
corresponds naturally to the identity shortcut mapping (especially
the pre-activation scheme in \cite{he2016identity}, which we adopted
in this paper).



\subsection{Formulation in terms of Dynamical Systems}
\label{sec:formal} 
We frame recurrent and residual neural networks in the language of
dynamical systems. We consider here dynamical systems in discrete
time, though most of the definitions carry over to continuous time.
A neural network (that we assume for simplicity 
to have a single layer with $n$ neurons) can be a dynamical system
with a dynamics defined as
\begin{equation}
h_{t+1}= f(h_t;w_t) + x_t
\end{equation}
where $h_t \in \R^n$ is the activity of the $n$ neurons in the layer
at time $t$ and $f:\R^n \to \R^n $ is a continuous, bounded function
parametrized by the vector of weights $w_t$. In a typical neural network,
$f$ is synthesized by the following relation between the activity $y_t$ of a
single neuron and its inputs $x_{t-1}$: 

\begin{equation}
y_t=\sigma(\scal{w}{x_{t-1}}+b),
\end{equation}

where $\sigma$ is a nonlinear function such as the linear rectifier
$\sigma(\cdot) = |\cdot|_+$.

A standard classification of dynamical systems defines the system as 

\begin{enumerate}
\item {\it homogeneous} if $x_t=0, \,\, \forall t > 0$ (alternatively 
  the equation reads as $h_{t+1}= f(h_t;w_t)$ with the inital
  condition $h_0=x_0$)
\item {\it time invariant} if $w_t=w$.
\end{enumerate}

Residual networks with weight sharing thus correspond to {\it
  homogeneous, time-invariant} systems which in turn correspond to a
feedback system (see Figure \ref{fig:feedback_system}) with an input
which is non-zero only at time $t=0$ ($x_{t=0} = x_0, x_t=0 \,\, \forall t > 0$) and with $f(z)=(K+I) \circ z$: 

\begin{equation}
 h_n = f(h_t;w_t) = (K+I)^n\circ x_0
\end{equation}

``Normal'' residual networks correspond to {\it homogeneous,
  time-variant} systems.  An analysis of the corresponding {\it
  inhomogeneous, time-invariant} system is provided in the
supplementary materials.

\section{A Generalized RNN for Multi-stage Fully Recurrent Processing}
As shown in the previous section, the recurrent form of a ResNet is
actually shallow (if we ignore the possible depth of the operator
$K$). In this section, we generalize it into a moderately deep RNN
that reflects the multi-stage processing in the primate visual cortex.

\begin{figure*}
  \renewcommand\figurename{ Figure}   
  \begin{center}
  \includegraphics[width=\linewidth]{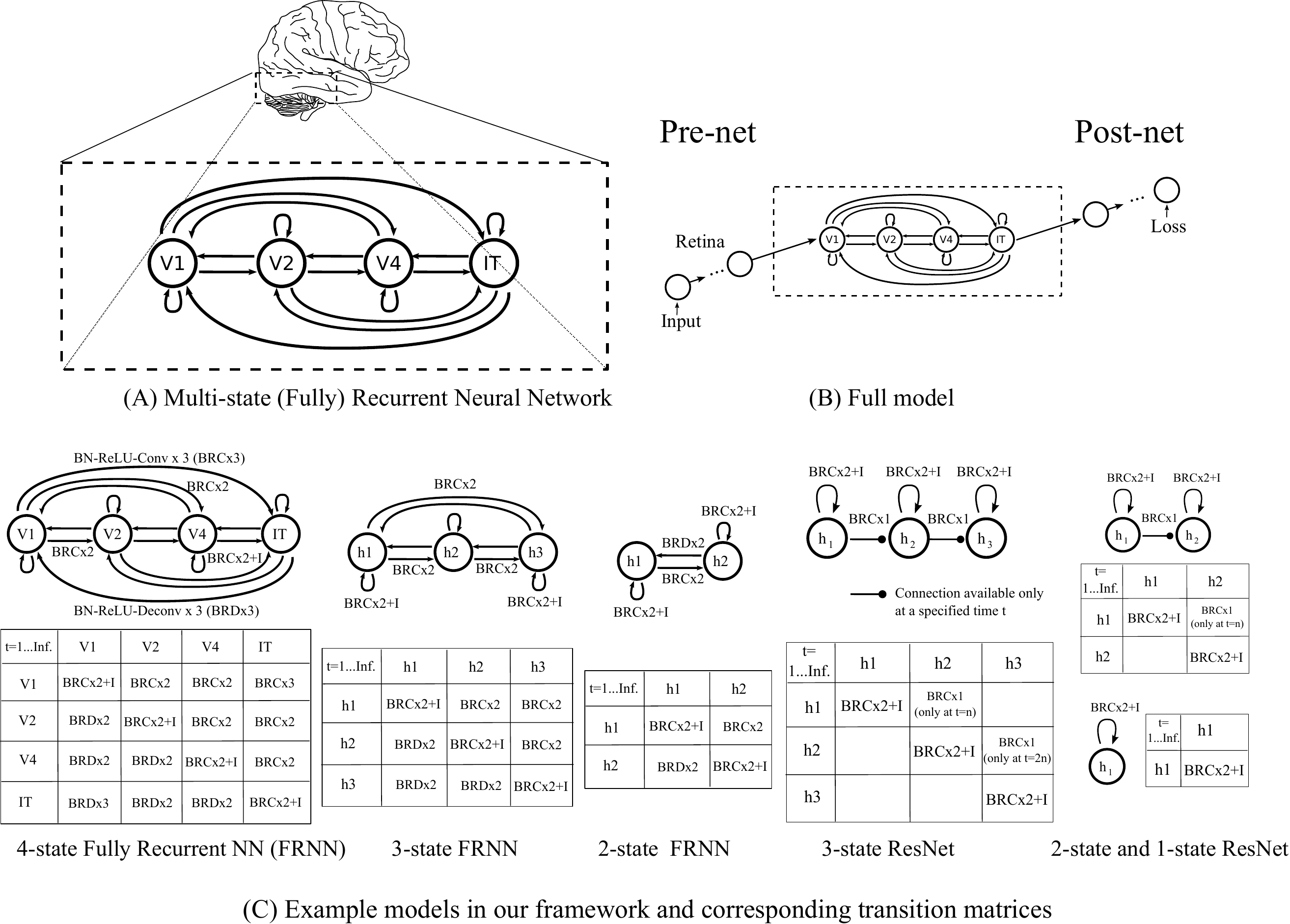}
  \end{center}
\caption{ (A) Modeling the ventral stream of visual cortex using
  a multi-state fully recurrent neural network. (B) the model consists
  of a input, a recurrent part and a output. (C) Examples of the
  recurrent part of the model and corresponding transition matrices
  used in the paper. ``BN'' denotes Batch Normalization and ``Conv''
  denotes convolution. Deconvolution layer (denoted by ``Deconv'') is
  \cite{zeiler2010deconvolutional} used as a transition function from
  a spacially small state to a spacially large one. BRCx2/BRDx2
  denotes a BN-ReLU-Conv/Deconv-BN-ReLU-Conv/Deconv pipeline (similar
  to a residual module \cite{he2016identity}). There is always a 2x2
  subsampling/upsampling between nearby states (e.g., V1/h1: 32x32,
  V2/h2: 16x16, V4/h3:8x8, IT:4x4). Stride 2 (convolution) or
  upsampling 2 (deconvolution) is used in transition functions to
  match the spacial sizes of input and output states. The intermediate
  feature sizes of transition function BRCx2/BRDx2 or BRCx3/BRDx3 are
  chosen to be the average feature size of input and output states.
  ``+I'' denotes a identity shortcut mapping. } 
\label{fig:main_model}
\end{figure*}

\textbf{Multi-state Graph}\label{sec:multi_state}  
We propose a general formulation that can capture the computations
performed by a multi-stage processing hierarchy with full recurrent
connections. Such a hierarchy can be characterized by a directed
(cyclic) graph G with vertices V and edges E: G = \{V,E\}. where 
vertices V is a set contains all the processing stages (i.e., we also
call them states). Take the ventral stream of visual cortex for
example, $V = \{LGN,V1,V2,V4,IT\}$. Note that retina is not listed
since there is no known feedback from primate cortex to the retina.
The edges $E$ are a set that contains all the connections (i.e.,
transition functions) between all vertices/states, e.g., V1-V2, V1-V4,
V2-IT, etc. One example of such a graph is in Figure
\ref{fig:main_model} (A).

  \renewcommand\figurename{ Figure} 
\begin{figure} 
\renewcommand\figurename{ Figure} 
  \centering  
  \includegraphics[width=\linewidth]{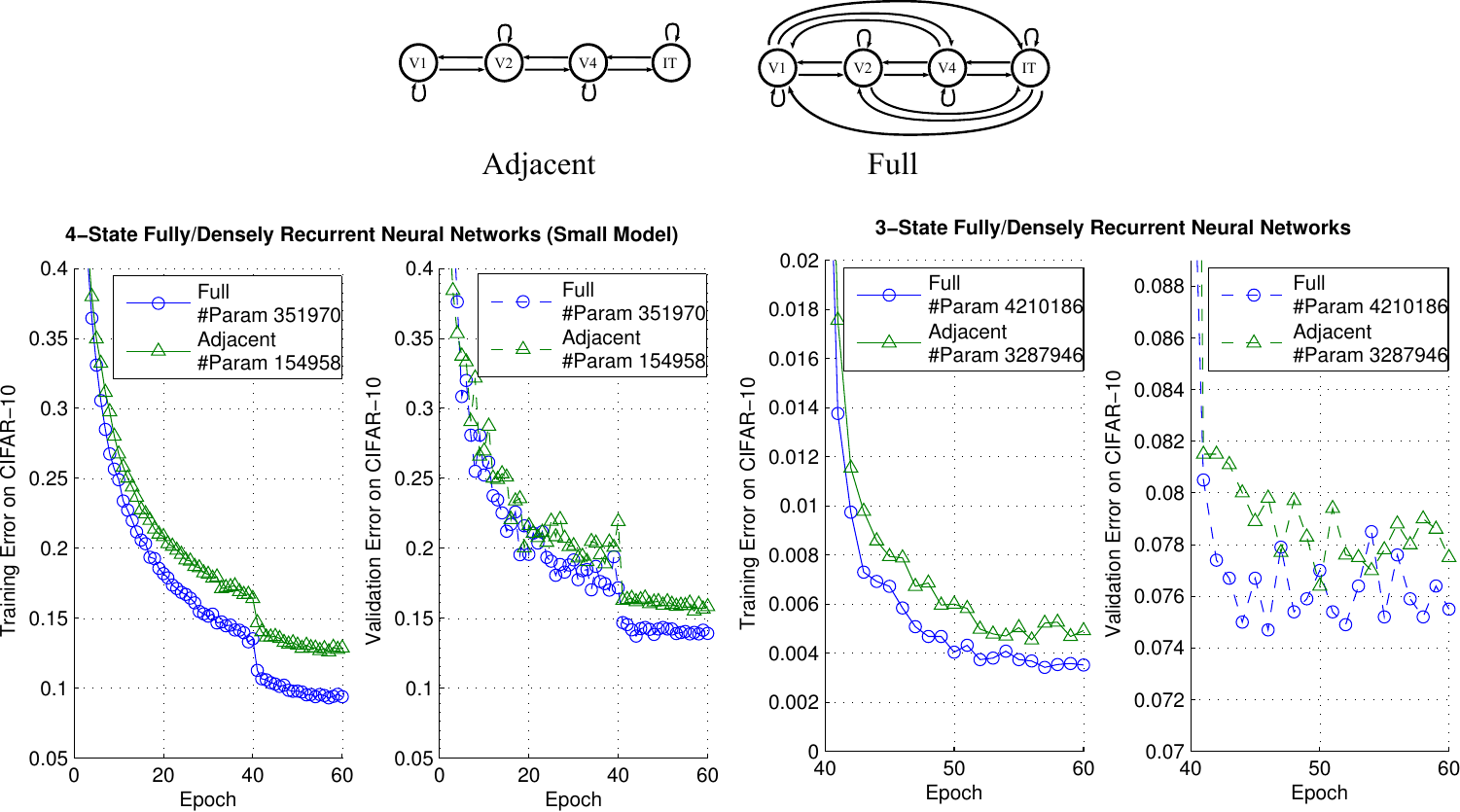}  
  
\caption{ The performance of 4-state and 3-state models. The state
  sizes of the 4-state model are: 32x32x8, 16x16x16, 8x8x32, 4x4x64.
  The state sizes of the 3-state model are: 32x32x64, 16x16x128,
  8x8x256. Only small 4-state models were tried since they are very 
  computationally heavy. The readout time is t=5 for both models. All
  models are time-invariant systems (i.e., weights are shared across
  time). } 
\label{fig:3states}
\end{figure} 


\begin{table}
\newcommand{\specialcell}[2][c]{%
\begin{tabular}[#1]{@{}c@{}}#2\end{tabular}}
 
  \centering   
  \captionsetup{width=.96\linewidth}
\begin{tabular}{  |c|c|c|c| }
 \hline
 Model  & Error (\%)                 & Depth  & \specialcell{Training \\ Epochs}    \\
 \hline
\specialcell{ All-CNN \\ \cite{springenberg2014striving} }  & 7.25\% & 11 & 350 \\ 
 \hline
\specialcell{ Highway Network \\ \cite{srivastava2015highway}} &  7.72\% & 19 & 400 \\
 \hline
\specialcell{ ResNet-110 \\ \cite{he2016identity}}  &  6.61\%         & 110  &  200  \\ 
 \hline
   ResNet-164  & 5.46\%                  & 164 &      200    \\
   ResNet-1001   & 4.69\%         & 1001  &   200   \\
   \cite{he2016identity} & & & \\ 
 \hline
 Human\cite{cifar10human}          & $\approx$ 6\%   & Recurrent  & -    \\
 \hline
\specialcell{ Same FRNN with \\ different readout time} & & Recurrent & \\ 
 3-state FRNN (readout t=5)  & 7.44\% &  13 (unrolled)  & 60 \\ 
 3-state FRNN (readout t=10)  & 6.86\% &  23 (unrolled)  & 60 \\
 \hline
\end{tabular}
  \renewcommand\tablename{ Table} 
  \captionof{table}{ Compare our model (FRNN) with other best models (prior to our work) on CIFAR-10. All models were trained with simple translation and mirror augmentation.  The depth is the number of weighted linear combination layers and pooling layers.  All of our models were trained with 60 epochs for consistency, since we did not focus on the absolute performance. We expect better performance if more epochs are used (will report in the next revision). Nevertheless, these models are all at the level of human performance. We believe at this point biological consistency is more interesting than marginal performance differences. Also our models achieve high performance with small latency (i.e., depth), which supports rapid visual recognition and is crucial for organism's survival.    }  
\label{table:performance}
\end{table} 


\textbf{Pre-net and Post-net}\label{sec:hbt}
The multi-state fully recurrent system does not have to receive raw
inputs. Rather, a (deep) neural network can serve as a preprocesser.
We call the preprocesser a ``pre-net'' as shown in Figure
\ref{fig:main_model} (B). On the other hand, one also needs a
``post-net'' as a postprocessor and provide supervisory signals to the
recurrent system and the pre-net. The pre-net, recurrent system and
post-net are trained in an end-to-end fashion with backpropagation.
 
For most models in this paper, unless stated otherwise, the pre-net is
a simple 3x3 convolutional layer and the post-net is a pipeline of a
batch normalization, a ReLU, a global average pooling and a fully
connected layer (or a 1x1 convolution, we use these terms
interchangeably).

Take primate visual system for instance, the retina is a part of the
``pre-net''. It does not receive any feedback from the cortex and thus
can be separated from the recurrent system for simplicity. In Section
\ref{sec:large_model}, we also tried 3 layers of 3x3 convolutions as
an pre-net, which might be more similar to a retina, and observed
slightly better performance.


\textbf{Transition Matrix} The connections between states over time
can be represented by a 3-D matrix where each element (i,j,t)
represents the {\it transition function} from state $i$ to state $j$
at time $t$. This formulation is flexible in the sense that transition
functions can vary over time (e.g., being blocked from time $t_1$ to
time $t_2$, etc.). This formulation allows us to design a system where
multiple locally recurrent systems are connected sequentially: a
downstream recurrent system only receives inputs when its upstream
recurrent system finishes, similar to recurrent convolutional neural
networks (e.g., \cite{liang2015recurrent}). This system with
non-shared weights can also represent exactly the state-of-the-art
ResNet (see supplementary materials for detailed illustrations).
Example transition matrices used in this paper are shown in Figure
\ref{fig:main_model} (C). When there are multiple transition functions
to a state, their outputs are summed (for ResNet) or averaged (for our
fully recurrent neural networks (FRNN) ). Averaging gives slightly
better performance for FRNN (about 1\%).

\textbf{Weight sharing} Given an unrolled network, a weight sharing
configuration can be described as a set $S$, whose element is a set of
tied weights $s=\{ W_{i_1,j_1,t_1},...,W_{i_m,j_m,t_m} \}$, where
$W_{i_m,j_m,t_m}$ denotes the weight of the transition functions from
state $i_m$ to $j_m$ at time $t_m$. This requires: 1. all weights
$W_{i_m,j_m,t_m} \in s$ have the same initial values. 2. the actual
gradients used for updating each element of $s$ is the sum of the
gradients of all elements in $s$: $ \forall W\in s, \,\,
(\frac{\partial E}{\partial W})_{used} = \sum_{W' \in s}
(\frac{\partial E}{\partial W'})_{original} $, where $E$ is the
training objective.

\textbf{Notations: Unrolling Depth vs. Readout Time} \label{sec:sim_time}
The meaning of ``unrolling depth'' may vary in different RNN models
since ``unrolling'' a cyclic graph is not well defined. In this paper,
we adopt a biologically-plausible definition: we simulate the time
after the onset of the visual stimuli assuming each transition
function takes constant time 1. We use the term ``readout time'' to
refer to the time the post-net reads the data from the last state.
Regarding the initial values, at $t=0$ all states are empty except
that the first state has some data received from the pre-net. We only
start simulate a transition function when its input state is
populated.

This definition in principle allows one to have quantitive comparisons
with biological systems. e.g., for a model with readout time 
$t$ in this paper, the wall clock time can be estimated to be $20t$ to
$50t$ ms, considering the latency of a single layer of biological
neurons.

\textbf{Sequential vs. Static Inputs/Outputs} \label{sec:seq_sta} As a
RNN, our model supports sequential data processing and in principle 
all other tasks supported by traditional RNNs. See supplementary
materials for illustrations and experiments on character-level
language modeling. 



\textbf{Batch Normalizations for RNNs} \label{sec:bn} As an additional
observation, we found that it generally hurts performance when the
normalization statistics (i.e., average, standard deviation) in batch
normalization are shared across time. This may explain the
difficulties observed in \cite{laurent2015batch}. However, good performance is restored if we apply a procedure we call
a ``time-specific normalization'': mean and standard deviation are
calculated independently for every $t$ (using training set). In
CIFAR-10 experiments (except Table \ref{table:performance}) we do not
use the learnable parameters of BN. In ImageNet experiments, we do
used the learnable parameters (shared over time) since they noticeably
improves performance.

\section{Related Work}
\label{sec:related_work}
\textbf{Deep Recurrent Neural Networks:} Our final model is deep and
similar to a stacked RNN
\cite{schmidhuber1992learning,el1995hierarchical,graves2013generating}
with several main differences: 1. our model has feedback transitions
between hidden layers and self-transition from each hidden layer to
itself. 2. our model has identity shortcut mappings inspired by
residual learning. 3. our transition functions are deep and
convolutional. As suggested by \cite{pascanu2013construct}, the term depth in RNN
could also refer to input-to-hidden, hidden-to-hidden or
hidden-to-output connections. Our model is deep in all of these
senses.  See Section \ref{sec:hbt}.

\textbf{Recursive Neural Networks and Convolutional Recurrent Neural
  Networks: } When unfolding RNN into a feedforward network, the
weights of many layers are tied. This is reminiscent of Recursive
Neural Networks (Recursive NN), first proposed by
\cite{socher2011parsing}. Recursive NN are characterized by applying
same operations recursively on a structure. The convolutional version
was first studied by \cite{eigen2013understanding}. Subsequent related
work includes \cite{pinheiro2013recurrent} and
\cite{liang2015recurrent}. One characteristic distinguishes our model
and residual learning from Recursive NN and convolutional recurrent NN
is whether there are identity shortcut mappings. This discrepancy
seems to account for the superior performance of residual learning and
of our model over the latters.

Our work may explain the seemingly surprising observation of
``stochastic depth'' \cite{huang2016deep} that randomly replacing a
subset of layers of ResNet by identity mappings during training does
not hurt performance --- if they are similar refinements performed
repeatedly, dropping a few of them should not be catastrophic. A
recent report \cite{caswellloopy} we became aware of after we finished
this work discusses the idea of imitating cortical feedback by
introducing loops into neural networks. A Highway
Network \cite{srivastava2015highway} is a feedforward network inspired
by Long Short Term Memory \cite{hochreiter1997long} featuring gated
shortcut mappings (instead of hardwired identity mappings used by
ResNet).

\section{Experiments}
\label{sec:exp_a} 
\textbf{Dataset and training details} We test most models on the
standard CIFAR-10 \cite{krizhevsky2009learning} dataset. All images
are 32x32 pixels with color. Data augmentation is performed in the
same way as \cite{he2015deep}. Momentum was used with hyperparameter
0.9. Experiments were run for 60 epochs with batchsize 64 unless
stated otherwise. The learning rates are 0.01 for the first 40 epochs,
0.001 for epoch 41 to 50 and 0.0001 for the last 10 epochs. All
experiments used the cross-entropy loss function and softmax for
classification. Batch Normalization (BN) \cite{ioffe2015batch} is used
for all experiments. But the learnable scaling and shifting parameters
are not used in the recurrent parts of the model unless stated otherwise.
Network weights were initialized with the method described in
\cite{he2015delving}. The implementations are based on MatConvNet\cite{vedaldi2015matconvnet}.  

\begin{figure}
  \renewcommand\figurename{ Figure }    
  \centering
\begin{center}

  \includegraphics[width=1\linewidth]{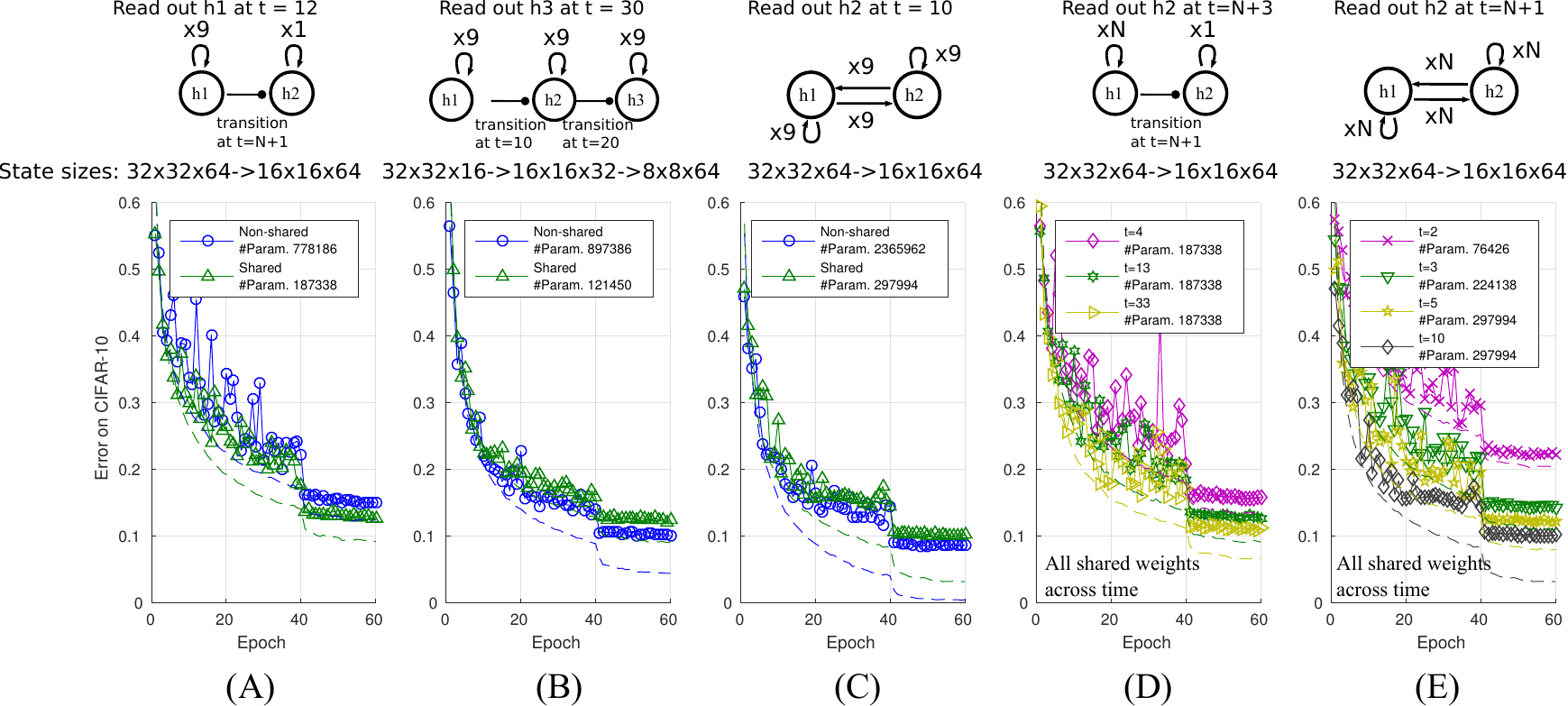}
\end{center}
\caption{ The transition matrices of all models are shown in Figure \ref{fig:main_model}C. ``\#Param'' denotes the number of parameters. For each model, the effective unrolling factors ``xN'' are determined by the readout time $t$ (See Section \ref{sec:sim_time} for the definition).  Dashed lines are training errors.  For multi-state ResNet, a downstream state only executes after receiving inputs from upstream states. For fully recurrent nets, all transitions execute concurrently. The state sizes are shown above. (A)  \textbf{A 2-state ResNet with the second state only unrolled once. This architecture always works better with shared rather than non-shared weights on CIFAR-10. In general, we empirically observe that weight sharing in the 1st state of a multi-state system is often beneficial (even in ImageNet).} One conjecture is that our transition function might match better low-level visual recurrent computations. (B) Standard 3-state ResNet. (C) 2-state fully recurrent NN. (D) Different readout time of (A). (E) A 2-state fully recurrent network with different readout time. There is consistent performance improvement as $t$ increases. The number of parameters changes since at some $t$, some recurrent connections have not been contributing to the output and thus their number of parameters are subtracted from the total. We observe in (D) and (E) that error reduces as t increases, while \#param. is kept the same.}  
\label{fig:share_vs_nonshare}
\end{figure}

\begin{figure} 
  \centering 
  \includegraphics[width=\linewidth]{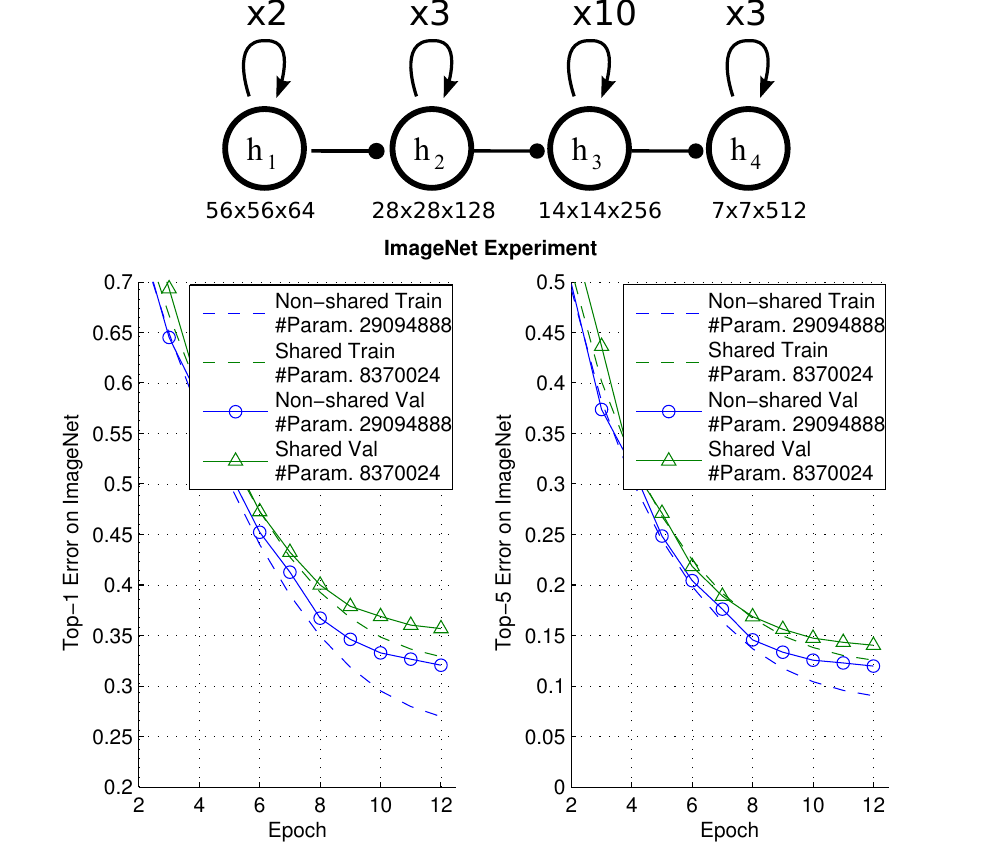}  
   
  \caption{ ImageNet experiment with a 4-state ResNet with non-shared vs. shared weights. Train: training error. Val: validation error.  No scaling augmentation is done. Performances are all based on single-crop. We trained the models with 12 epochs and logarithmically decaying learning rates (i.e., logspace(-1, -4, N) in Matlab, where N=12 is the number of epochs). The logarithmic learning rates make the model converge much faster than the usual three stage learning rates (i.e., dividing by 10 every some epochs), thus the models have already converged well and the performance differences are representative. }   
  \normalsize
  \label{fig:imagenet}
\end{figure}

\textbf{ResNet with Shared Weights Across Time}
We conjecture that the effectiveness of ResNet mainly comes from the fact
that it efficiently models the recurrent computations required by the
recognition task. If this is the case, one should be able to
reinterpret ResNet as a RNN with weight sharing and achieve comparable
performance to the original version. We demonstrate various
incarnations of this idea and show it is indeed the case (See Figure \ref{fig:share_vs_nonshare} and Figure \ref{fig:imagenet}).

\textbf{Fully Recurrent Neural Networks with Shared and Non-shared Weights}
Although a RNN is usually implemented with shared weights across
time, it is however possible to unshare the weights and use an
independent set of weights at every time $t$. For practical
applications, whenever one can have a initial $t=0$ and enumerate all
possible $t$s, a RNN with non-shared weights should be feasible,
similar to the time-specific batch normalization described in
\ref{sec:bn}. The results of 2-state fully recurrent neural networks 
with shared and non-shared weights are shown in Figure \ref{fig:share_vs_nonshare} (C). 
 


\begin{figure} 
  \renewcommand\figurename{ Figure} 
\centering
  \centering
  \includegraphics[width=\linewidth]{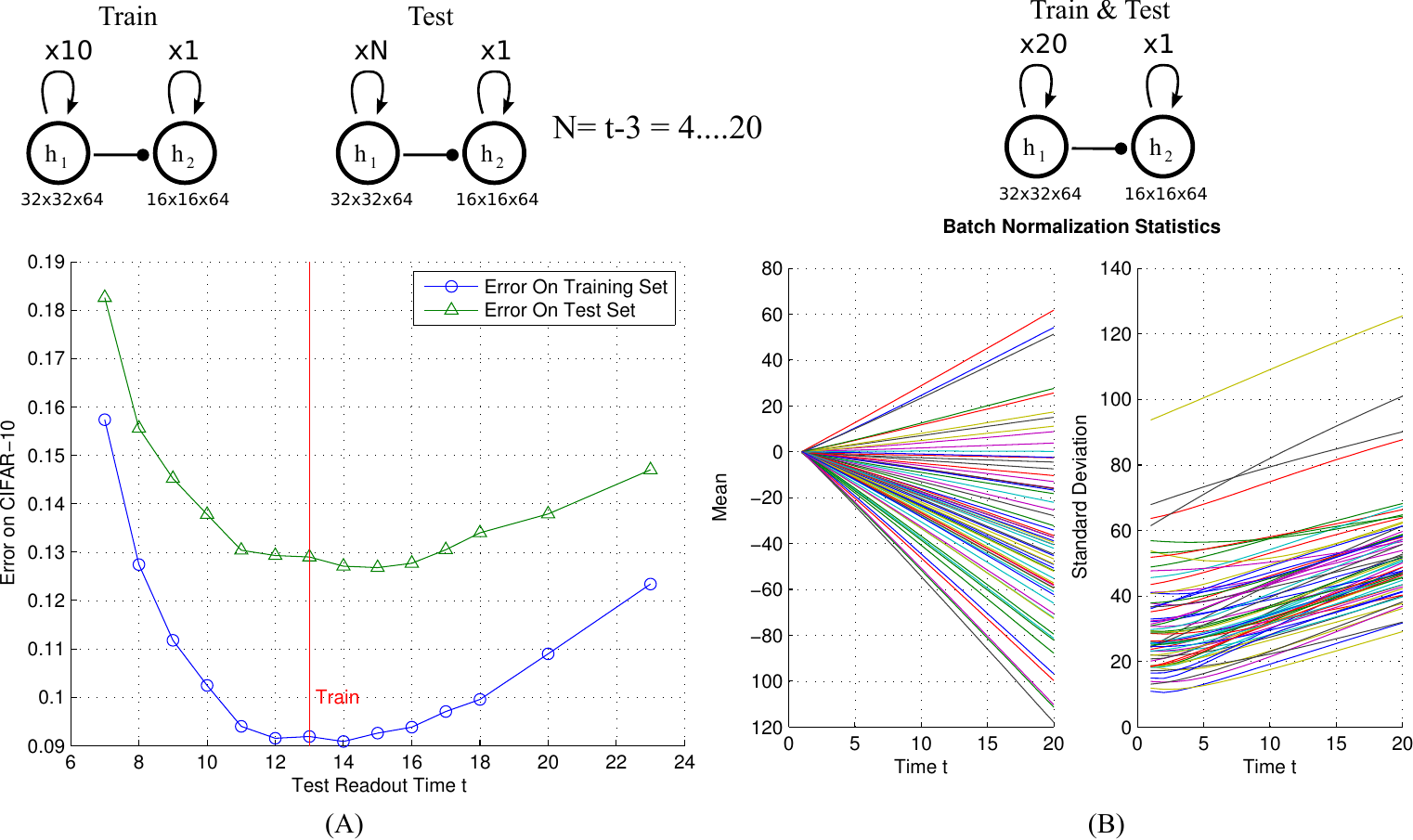} 
  
  \caption{ (A) Training and testing with different readout time. A 2-state ResNet with shared weights is trained with readout time t=13 (i.e., 1st state unrolling factor N=t-3=10) and tested with t from 7 to 23.  The model can generalize well to readout time that are not trained with. There even seems to be tiny performance improvement (on test set) when testing with slightly larger readout time than training. (B) Biological Plausibility of Time-specific Batch Normalization (TSBN): we show the normalization statistics of a batch normalization module in the model in part (B) (which was trained with t=23). Each line represents the statistics of a feature channel over time. TSBN seems to implement a simple decay, thus it may be biologically-plausible.    } 
  \normalsize
  \label{fig:gen_readout} 
\end{figure}

\textbf{The Effect of Readout Time} In visual cortex, useful
information increases as time proceeds from the onset of the visual
stimuli. This suggests that recurrent system might have better
representational power as more time is allowed. We tried training and
testing 2-state ResNet and FRNN with various
readout time (i.e., unrolling depth, see Section \ref{sec:sim_time})
and observe similar effects. Results are shown in Figure \ref{fig:share_vs_nonshare}
(D) and (E). The difference from Figure \ref{fig:gen_readout} (A) is that  
training and testing readout times are the same in this experiment. 


\textbf{Larger Models With More States}
\label{sec:large_model}

The results of 3-state and 4-state FRNNs are shown in
Figure \ref{fig:3states} and Table \ref{table:performance}.
3-state models seem to generally outperform 2-state ones. This is
expected since more parameters are introduced. We compare the 3-state
models with existing best models on CIFAR-10 in
Table \ref{table:performance}.Next, for computational efficiency, we
tried only allowing each state to have transitions to adjacent states
and to itself by disabling bypass connections (e.g., Figure \ref{fig:3states} top). In this case, the number of transitions scales linearly as the
number of states increases, instead of quadratically. This setting
performs well with 3-state networks and slightly less well with
4-state networks (perhaps as a result of small feature/parameter
sizes). Finally, for 4-state fully recurrent networks, the models tend
to become overly computationally heavy if we train it with large $t$
or large number of feature maps. With small $t$ and feature maps, we
have not achieved better performance than 3-state networks. For experiments in this
subsection, we choose a moderately deep pre-net of three 3x3
convolutional layers to model the layers between retina and V1:
Conv-BN-ReLU-Conv-BN-ReLU-Conv. This is not essential but outperforms
shallow pre-net slightly (within 1\% validation error).



\textbf{Generalization Across Readout Time} As a RNN, our model supports training and testing with different readout times. Based on our theoretical analyses in Section \ref{sec:formal}, the representation is usually not guaranteed to converge when running a model with time $t \to \infty$. Nevertheless, the model exhibits good generalization over time. Results are shown in Figure \ref{fig:gen_readout} and supplementary materials.

\textbf{Experiments on ImageNet} We also evaluated weight sharing on ImageNet 1000-way classification task. The results are shown in Figure \ref{fig:imagenet}.


\section{Discussion}

\textbf{When is Weight Sharing Good/Bad?} We observe that (e.g., in Figure \ref{fig:share_vs_nonshare} A),
sharing weights over time in the first stage of ResNet always improves
performance. However, sharing weights in the later stages does
sometimes slightly lower performance. We conjecture that whenever our
prior --- the transition function (e.g., BN-ReLU-Conv x2) is
``similar'' to the underlying recurrent function used by visual
cortex, weight sharing will be beneficial. This would imply that our
transition function happen to better model recurrent computations in early
visual areas than later stages. 

\textbf{Psychophysics Support:} 
After our work, \cite{eberhardt2016deep} reported that while higher
 layers of deep networks enjoy better performance on a visual
 recognition task, only intermediate (instead of last) layers agree
 best with human rapid predictions (t= 50 to 500ms). Given that a deep
 network's output layer is trained to match human predictions at
 $t=\infty$. It is interesting to see an intermediate layer matches best
 an intermediate time, consistent with what we predict.

\textbf{ Future Directions:} \textbf{1. The dark secret of Deep Networks: trying to imitate Recurrent Shallow Networks?}
The results of this paper lead to the following conjecture: the 
effectiveness of most of the deep feedforward neural networks,
including but not limited to ResNet, can be attributed to their
ability to approximate recurrent computations that are prevalent in
most tasks with larger $t$ than shallow feedforward networks. This may
offer a new perspective on the theoretical pursuit of the
long-standing question ``why is deep better than shallow''
\cite{montufar2014number,mhaskar2016learning}. \textbf{2. Conjecture about Cortex and Recurrent Computations in Cortical Areas:}
Most of the models of cortex that led to the Deep Convolutional
architectures and followed them -- such as the Neocognitron
\cite{Fukushima1980}, HMAX \cite{Riesenhuber1999} and more recent
models \cite{YaminsDicarlo2016} -- have neglected the layering in each
cortical area and the feedforward and recurrent connections within
each area and between them. They also neglected the time evolution of
selectivity and invariance in each of the areas. Our proposal makes
several interesting predictions. Each cortical area would correspond
to a recurrent network and thus to a system with a temporal dynamics
even for flashed inputs; with increasing time one expects
asymptotically better performance; masking with a mask an input image
flashed briefly should disrupt recurrent computations in each area;
performance should increase with time even without a mask for briefly
flashed images; the cortex and each of its component areas are RNNs
and, unlike relatively shallow feedforward networks, are computationally as
powerful as universal Turing machine \cite{siegelmann1992computational}.

\bibliographystyle{apalike} 

\bibliography{rsdn}

\begin{thebibliography}{}

\bibitem[B{\"u}chel and Friston, 1997]{buchel1997modulation}
B{\"u}chel, C. and Friston, K. (1997).
\newblock Modulation of connectivity in visual pathways by attention: cortical
  interactions evaluated with structural equation modelling and fmri.
\newblock {\em Cerebral cortex}, 7(8):768--778.

\bibitem[Caswell et~al., 2016]{caswellloopy}
Caswell, I., Shen, C., and Wang, L. (2016,
  \url{http://cs231n.stanford.edu/reports2016/110_Report.pdf}. Google Scholar
  time stamp: March 25th, 2016).
\newblock Loopy neural nets: Imitating feedback loops in the human brain.
\newblock {\em CS231n Report, Stanford}.

\bibitem[Eberhardt et~al., 2016]{eberhardt2016deep}
Eberhardt, S., Cader, J., and Serre, T. (2016).
\newblock How deep is the feature analysis underlying rapid visual
  categorization?
\newblock {\em arXiv preprint arXiv:1606.01167}.

\bibitem[Eigen et~al., 2013]{eigen2013understanding}
Eigen, D., Rolfe, J., Fergus, R., and LeCun, Y. (2013).
\newblock Understanding deep architectures using a recursive convolutional
  network.
\newblock {\em arXiv preprint arXiv:1312.1847}.

\bibitem[El~Hihi and Bengio, ]{el1995hierarchical}
El~Hihi, S. and Bengio, Y.
\newblock Hierarchical recurrent neural networks for long-term dependencies.
\newblock Citeseer.

\bibitem[Fukushima, 1980]{Fukushima1980}
Fukushima, K. (1980).
\newblock {Neocognitron: A self-organizing neural network model for a mechanism
  of pattern recognition unaffected by shift in position}.
\newblock {\em Biological Cybernetics}, 36(4):193--202.

\bibitem[Graves, 2013]{graves2013generating}
Graves, A. (2013).
\newblock Generating sequences with recurrent neural networks.
\newblock {\em arXiv preprint arXiv:1308.0850}.

\bibitem[He et~al., 2015a]{he2015deep}
He, K., Zhang, X., Ren, S., and Sun, J. (2015a).
\newblock Deep residual learning for image recognition.
\newblock {\em arXiv preprint arXiv:1512.03385}.

\bibitem[He et~al., 2015b]{he2015delving}
He, K., Zhang, X., Ren, S., and Sun, J. (2015b).
\newblock Delving deep into rectifiers: Surpassing human-level performance on
  imagenet classification.
\newblock In {\em Proceedings of the IEEE International Conference on Computer
  Vision}, pages 1026--1034.

\bibitem[He et~al., 2016]{he2016identity}
He, K., Zhang, X., Ren, S., and Sun, J. (2016).
\newblock Identity mappings in deep residual networks.
\newblock {\em arXiv preprint arXiv:1603.05027}.

\bibitem[Hochreiter and Schmidhuber, 1997]{hochreiter1997long}
Hochreiter, S. and Schmidhuber, J. (1997).
\newblock Long short-term memory.
\newblock {\em Neural computation}, 9(8):1735--1780.

\bibitem[Huang et~al., 2016]{huang2016deep}
Huang, G., Sun, Y., Liu, Z., Sedra, D., and Weinberger, K. (2016).
\newblock Deep networks with stochastic depth.
\newblock {\em arXiv preprint arXiv:1603.09382}.

\bibitem[Hupe et~al., 1998]{hupe1998cortical}
Hupe, J., James, A., Payne, B., Lomber, S., Girard, P., and Bullier, J. (1998).
\newblock Cortical feedback improves discrimination between figure and
  background by v1, v2 and v3 neurons.
\newblock {\em Nature}, 394(6695):784--787.

\bibitem[Ioffe and Szegedy, 2015]{ioffe2015batch}
Ioffe, S. and Szegedy, C. (2015).
\newblock Batch normalization: Accelerating deep network training by reducing
  internal covariate shift.
\newblock {\em arXiv preprint arXiv:1502.03167}.

\bibitem[Ito and Gilbert, 1999]{ito1999attention}
Ito, M. and Gilbert, C.~D. (1999).
\newblock Attention modulates contextual influences in the primary visual
  cortex of alert monkeys.
\newblock {\em Neuron}, 22(3):593--604.

\bibitem[Karpathy, 2011]{cifar10human}
Karpathy, A. (2011).
\newblock Lessons learned from manually classifying cifar-10.

\bibitem[Krizhevsky, 2009]{krizhevsky2009learning}
Krizhevsky, A. (2009).
\newblock Learning multiple layers of features from tiny images.

\bibitem[Krizhevsky et~al., 2012]{krizhevsky2012imagenet}
Krizhevsky, A., Sutskever, I., and Hinton, G.~E. (2012).
\newblock Imagenet classification with deep convolutional neural networks.
\newblock In {\em Advances in neural information processing systems}, pages
  1097--1105.

\bibitem[Lamme et~al., 1998]{lamme1998feedforward}
Lamme, V.~A., Super, H., and Spekreijse, H. (1998).
\newblock Feedforward, horizontal, and feedback processing in the visual
  cortex.
\newblock {\em Current opinion in neurobiology}, 8(4):529--535.

\bibitem[Laurent et~al., 2015]{laurent2015batch}
Laurent, C., Pereyra, G., Brakel, P., Zhang, Y., and Bengio, Y. (2015).
\newblock Batch normalized recurrent neural networks.
\newblock {\em arXiv preprint arXiv:1510.01378}.

\bibitem[Le et~al., 2015]{le2015simple}
Le, Q.~V., Jaitly, N., and Hinton, G.~E. (2015).
\newblock A simple way to initialize recurrent networks of rectified linear
  units.
\newblock {\em arXiv preprint arXiv:1504.00941}.

\bibitem[Liang and Hu, 2015]{liang2015recurrent}
Liang, M. and Hu, X. (2015).
\newblock Recurrent convolutional neural network for object recognition.
\newblock In {\em Proceedings of the IEEE Conference on Computer Vision and
  Pattern Recognition}, pages 3367--3375.

\bibitem[Liao et~al., 2015]{liao2015important}
Liao, Q., Leibo, J.~Z., and Poggio, T. (2015).
\newblock How important is weight symmetry in backpropagation?
\newblock {\em arXiv preprint arXiv:1510.05067}.

\bibitem[Mhaskar et~al., 2016]{mhaskar2016learning}
Mhaskar, H., Liao, Q., and Poggio, T. (2016).
\newblock Learning real and boolean functions: When is deep better than
  shallow.
\newblock {\em arXiv preprint arXiv:1603.00988}.

\bibitem[Montufar et~al., 2014]{montufar2014number}
Montufar, G.~F., Pascanu, R., Cho, K., and Bengio, Y. (2014).
\newblock On the number of linear regions of deep neural networks.
\newblock In {\em Advances in neural information processing systems}, pages
  2924--2932.

\bibitem[Pascanu et~al., 2013]{pascanu2013construct}
Pascanu, R., Gulcehre, C., Cho, K., and Bengio, Y. (2013).
\newblock How to construct deep recurrent neural networks.
\newblock {\em arXiv preprint arXiv:1312.6026}.

\bibitem[Pinheiro and Collobert, 2013]{pinheiro2013recurrent}
Pinheiro, P.~H. and Collobert, R. (2013).
\newblock Recurrent convolutional neural networks for scene parsing.
\newblock {\em arXiv preprint arXiv:1306.2795}.

\bibitem[Rao and Ballard, 1999]{rao1999predictive}
Rao, R.~P. and Ballard, D.~H. (1999).
\newblock Predictive coding in the visual cortex: a functional interpretation
  of some extra-classical receptive-field effects.
\newblock {\em Nature neuroscience}, 2(1):79--87.

\bibitem[Riesenhuber and Poggio, 1999]{Riesenhuber1999}
Riesenhuber, M. and Poggio, T. (1999).
\newblock {Hierarchical models of object recognition in cortex}.
\newblock {\em Nature Neuroscience}, 2(11):1019--1025.

\bibitem[Schmidhuber, 1992]{schmidhuber1992learning}
Schmidhuber, J. (1992).
\newblock Learning complex, extended sequences using the principle of history
  compression.
\newblock {\em Neural Computation}, 4(2):234--242.

\bibitem[Serre et~al., 2007]{Serre2007}
Serre, T., Oliva, A., and Poggio, T. (2007).
\newblock {A feedforward architecture accounts for rapid categorization}.
\newblock {\em Proceedings of the National Academy of Sciences of the United
  States of America}, 104(15):6424--6429.

\bibitem[Siegelmann and Sontag, 1992]{siegelmann1992computational}
Siegelmann, H.~T. and Sontag, E.~D. (1992).
\newblock On the computational power of neural nets.
\newblock In {\em Proceedings of the fifth annual workshop on Computational
  learning theory}, pages 440--449. ACM.

\bibitem[Socher et~al., 2011]{socher2011parsing}
Socher, R., Lin, C.~C., Manning, C., and Ng, A.~Y. (2011).
\newblock Parsing natural scenes and natural language with recursive neural
  networks.
\newblock In {\em Proceedings of the 28th international conference on machine
  learning (ICML-11)}, pages 129--136.

\bibitem[Springenberg et~al., 2014]{springenberg2014striving}
Springenberg, J.~T., Dosovitskiy, A., Brox, T., and Riedmiller, M. (2014).
\newblock Striving for simplicity: The all convolutional net.
\newblock {\em arXiv preprint arXiv:1412.6806}.

\bibitem[Srivastava et~al., 2015]{srivastava2015highway}
Srivastava, R.~K., Greff, K., and Schmidhuber, J. (2015).
\newblock Highway networks.
\newblock {\em arXiv preprint arXiv:1505.00387}.

\bibitem[Thorpe et~al., 1996]{thorpe1996speed}
Thorpe, S., Fize, D., Marlot, C., et~al. (1996).
\newblock Speed of processing in the human visual system.
\newblock {\em nature}, 381(6582):520--522.

\bibitem[Vedaldi and Lenc, 2015]{vedaldi2015matconvnet}
Vedaldi, A. and Lenc, K. (2015).
\newblock Matconvnet: Convolutional neural networks for matlab.
\newblock In {\em Proceedings of the 23rd Annual ACM Conference on Multimedia
  Conference}, pages 689--692. ACM.

\bibitem[Yamins and Dicarlo, 2016]{YaminsDicarlo2016}
Yamins, D. and Dicarlo, J. (2016).
\newblock Using goal-driven deep learning models to understand sensory cortex.

\bibitem[Zeiler et~al., 2010]{zeiler2010deconvolutional}
Zeiler, M.~D., Krishnan, D., Taylor, G.~W., and Fergus, R. (2010).
\newblock Deconvolutional networks.
\newblock In {\em Computer Vision and Pattern Recognition (CVPR), 2010 IEEE
  Conference on}, pages 2528--2535. IEEE.

\end{thebibliography}
\normalsize

\end{document}